\pdfoutput=1

\documentclass[11pt]{article}

\usepackage[final]{acl}

\usepackage{times}
\usepackage{latexsym}

\usepackage[T1]{fontenc}

\usepackage[utf8]{inputenc}

\usepackage{microtype}

\usepackage{inconsolata}

\usepackage{color,xcolor,colortbl}
\usepackage{booktabs}
\usepackage{diagbox}
\usepackage{graphicx}
\usepackage{xspace}
\usepackage{multirow}
\usepackage{amsmath}
\usepackage{amssymb}

\usepackage{algorithm}
\usepackage{algorithmicx}
\usepackage[noend]{algpseudocode}

\usepackage{tikz}
\usepackage{tikz-qtree}
\usepackage{pgfplots}
\usepackage{subfigure}

\usepackage{pgfplots}
\usepackage{pgfplotstable}

\definecolor{mydarkred}{RGB}{139, 0, 0}
\definecolor{blank}{HTML}{E6E6E6}

\definecolor{darkgreen}{RGB}{0, 200, 0}
\newcommand{\tool}{ReVD\xspace}
\newcommand{\moduleA}{BVD\xspace}
\newcommand{\moduleB}{T-SFT\xspace}
\newcommand{\moduleC}{COPO\xspace}
\newcommand{\piref}{\pi_\text{ref}}

\newcommand{\http}{\url{https://github.com/Xin-Cheng-Wen/PO4Vul}}

\title{Boosting Vulnerability Detection of LLMs via Curriculum Preference Optimization with Synthetic Reasoning Data}

\author{
 \textbf{Xin-Cheng Wen\textsuperscript{1}},\quad
 \textbf{Yijun Yang\textsuperscript{1}},\quad
 \textbf{Cuiyun Gao \textsuperscript{2}}\thanks{~~Corresponding author.},\quad
 \textbf{Yang Xiao\textsuperscript{3}},\quad
 \textbf{Deheng Ye\textsuperscript{1}}
\\
 \textsuperscript{1} Tencent Inc., China \\
 \textsuperscript{2} The Chinese University of Hong Kong, China\\
  \textsuperscript{3} Chinese Academy of Sciences, China \\
 \small{
   {xiamenwxc@foxmail.com, yijun.steven.yang@gmail.com, cygao@cse.cuhk.edu.hk}
 }\\
}

\begin{document}
\maketitle
\begin{abstract}

Large language models (LLMs) demonstrate considerable proficiency in numerous coding-related tasks; however, their capabilities in detecting software vulnerabilities remain limited. This limitation primarily stems from two factors: (1) the absence of reasoning data related to vulnerabilities, which hinders the models' ability to capture 
underlying vulnerability patterns; and (2) their focus on learning semantic representations rather than the reason behind them, thus failing to recognize semantically similar vulnerability samples. Furthermore, the development of LLMs specialized in vulnerability detection is challenging, particularly in environments characterized by the scarcity of high-quality datasets.
In this paper, we propose a novel framework \tool that excels at mining vulnerability patterns through reasoning data synthesizing and vulnerability-specific preference optimization. Specifically, we construct forward and backward reasoning processes for vulnerability and corresponding fixed code, ensuring the synthesis of high-quality reasoning data. Moreover, we design the triplet supervised fine-tuning followed by curriculum online preference optimization for enabling \tool to better understand vulnerability patterns.
The extensive experiments conducted on PrimeVul and SVEN datasets demonstrate that \tool sets new state-of-the-art for LLM-based software vulnerability detection, e.g., 12.24\%-22.77\% improvement in the accuracy. The source code and data are available at \http.

\end{abstract}

\section{Introduction}\label{sec:intro} 
Software vulnerabilities, primarily due to insecure coding, pose significant risks as they can be exploited to compromise software systems, leading to severe security issues~\cite{DBLP:conf/kbse/WenWGWLG23, DBLP:journals/tse/WenGLWLL24}.
Thousands of software vulnerabilities are discovered per year~\cite{Statista1}, underscoring the critical importance of vulnerability detection methods~\cite{devign,reveal,DBLP:conf/icse/CaoSBWLT22/mvd}. In recent years, Code Pre-Trained Models (CodePTMs) such as CodeBERT~\cite{DBLP:conf/emnlp/FengGTDFGS0LJZ20/codebert} and UniXcoder~\cite{DBLP:conf/acl/GuoLDW0022/unixcoder} have proven to be highly effective in identifying software vulnerabilities. They leverage extensive pre-training on diverse codebases to enhance their understanding and detection capabilities~\cite{CodeGeeX}. Following this paradigm, Large Language Models (LLMs) pre-trained on code data excel in comprehending and interpreting the semantics of both human and programming languages~\cite{DBLP:journals/corr/abs-2308-10620/se}, offering superior intelligence and flexibility for coding-related tasks. Despite the substantial advancements of LLMs, their performance in vulnerability detection task is still limited mainly due to the following two aspects: \looseness-1

\begin{figure*}[t]
	\centering
	\includegraphics[width=0.95\textwidth]{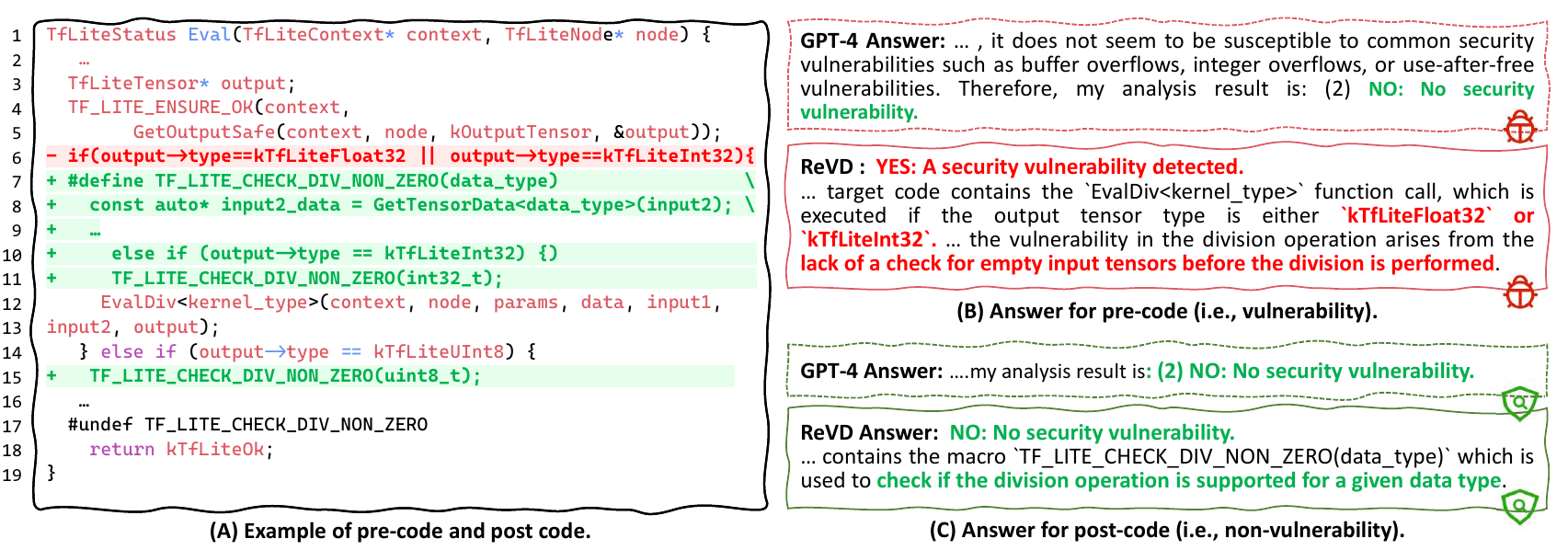}
    \caption{ 
\textbf{The vulnerability example of ``Divide By Zero'' (i.e.,~\cite{CVE-2021-37683}).}  \textit{Figure (A):} The \textcolor{red}{pre-code} (i.e., labeled as \textcolor{red}{vulnerable}) and \textcolor{darkgreen}{post-code} (i.e., labeled as \textcolor{darkgreen}{non-vulnerable}). The code shaded in \textcolor{red}{red} and \textcolor{darkgreen}{green} denote the \textcolor{red}{vulnerable} and \textcolor{darkgreen}{fixed} code, respectively. \textit{Figure (B):} GPT-4 and \tool's answer for \textcolor{red}{pre-code}. \textit{Figure (C):} GPT-4 and \tool's answer for \textcolor{darkgreen}{post-code}.} 
\vspace{-3pt}
\label{example}
\end{figure*}


\textbf{(1) The absence of reasoning data behind vulnerability-fixing processes hinders the model from understanding and aligning with established vulnerability patterns.} 
The current approaches~\cite{DBLP:journals/corr/abs-2310-09810/llm4vul, DBLP:conf/msr/FuT22/linevul} solely utilize extensive pre-training across diverse codebases 
to enhance their vulnerability detection capabilities, yet acquiring specific reasoning knowledge about vulnerability-fixing processes remains challenging. 
These methods do not explicitly generate step-by-step reasoning data related to vulnerabilities for model training,
and the thought processes of the individuals who fix vulnerabilities are not usually reflected in the dataset annotations.
As illustrated in Figure~\ref{example} (A), we present an example of a Divide By Zero vulnerability~\cite{CWE369}, where the code attempts to ``\texttt{output->type == kTfLiteInt32}''.
However, this may inadvertently lead to a division by zero within the division operations in ``\texttt{EvalDiv}''.
Therefore, the subsequent post-code employs the ``\texttt{TF\_LITE\_CHECK\_DIV\_NON\_ZERO}''. This macro effectively checks for and prevents division by zero errors in division operations, thereby also averting potential divide by zero vulnerability.
It is evident that existing vulnerability data often includes pre- and post-code snippets and lacks comprehensive reasoning data necessary to understand the triggers of these vulnerabilities.
Hence, GPT-4 fails to detect this vulnerability in Figure~\ref{example} (B).
\textbf{(2) Existing methods tend to 
focus on learning semantic representations rather than vulnerability patterns, which 
fail to distinguish the vulnerability from the fixed one due to the highly semantic similarity of the code.}
Vulnerability patches often involve subtle code changes such as adjusting buffer sizes, correcting data types, or adding security checks~\cite{DBLP:conf/icse/Luo0024}. These modifications typically result in a post-code (i.e., non-vulnerable) that is semantically similar to their pre-code (i.e., vulnerable). 
For instance, as shown in Figure~\ref{example} (B) and (C), GPT-4 does not discern the subtle differences between the vulnerability and the corresponding fixed code and generates the same answer for them. More specifically, GPT-4 fails to distinguish 78.62\% pairs of vulnerabilities and corresponding fixed code in realistic vulnerability detection scenarios from the PrimeVul~\cite{DBLP:journals/corr/abs-2403-18624/primevul} dataset. \looseness-1


To mitigate the above issues, in this paper, we propose a novel framework \tool that excels at mining vulnerability patterns through
vulnerability-specific preference optimization with reasoning-centric data synthesis. Specifically, we purposefully design three modules: \textbf{(1) Bi-directional Vulnerability Data Generation (\moduleA).} It consists of forward and backward reasoning processes for vulnerability and corresponding fixed code and utilizes the vulnerability information to generate a high-quality vulnerability reasoning dataset with 28k samples.
\textbf{(2) Triplet Supervised Fine-Tuning (\moduleB).} We propose a novel triplet loss function by analyzing the relationships among pre-code, post-code, and code-diff.
Fine-tuning LLMs with such a loss enhances their ability to capture vulnerability patterns as it enables consistency checks between their forward and backward reasoning.
\textbf{(3) Curriculum Online Preference Optimization (\moduleC).} The LLM focuses on learning the vulnerability patterns on which it underperforms via iterative preference optimization with instance- and task-level curricula, continuously enhancing its effectiveness
in real-world complex scenarios.

We evaluate \tool and compare it with nine representative CodePTMs and LLM-based vulnerability detection baselines. The extensive evaluations highlight that \tool outperforms all baselines and sets new state-of-the-art across two widely-used datasets PrimeVul~\cite{DBLP:journals/corr/abs-2403-18624/primevul} and SVEN~\cite{DBLP:conf/ccs/HeV23/SVEN}, with improvements of 9.08\%, 10.33\%, and 18.15\% in terms of accuracy, F1-Score, and VP-Score, respectively. More importantly, we also release the first vulnerability reasoning
dataset and the corresponding preference dataset, shedding novel insights on training more generalizable and versatile LLM experts for vulnerability detection in real-world scenarios.
\textbf{The major contributions of this paper are summarized below:}

\textbf{Reasoning Data:} 
We draw inspiration from the surge of large reasoning models~\cite{guo2025deepseek,cai2024system,cai2025mm}, find the substantial benefits of reasoning data for vulnerability detection, and propose the first fully automated pipeline to synthesize vulnerability reasoning data.

\textbf{Method:} 
We develop a novel two-stage framework \tool on our synthetic reasoning data, which can turn open-source LLMs into strong vulnerability detectors by aligning their coding knowledge with vulnerability semantics, patterns, and types.

\textbf{Results:} 
The results demonstrate that \tool sets new state-of-the-art for LLM-based software vulnerability detection, e.g., 12.24\%-22.77\% improvement in the accuracy.

\section{Related Work}\label{sec:related}

\begin{figure*}[t]
	\centering
	\includegraphics[width=0.95\textwidth]{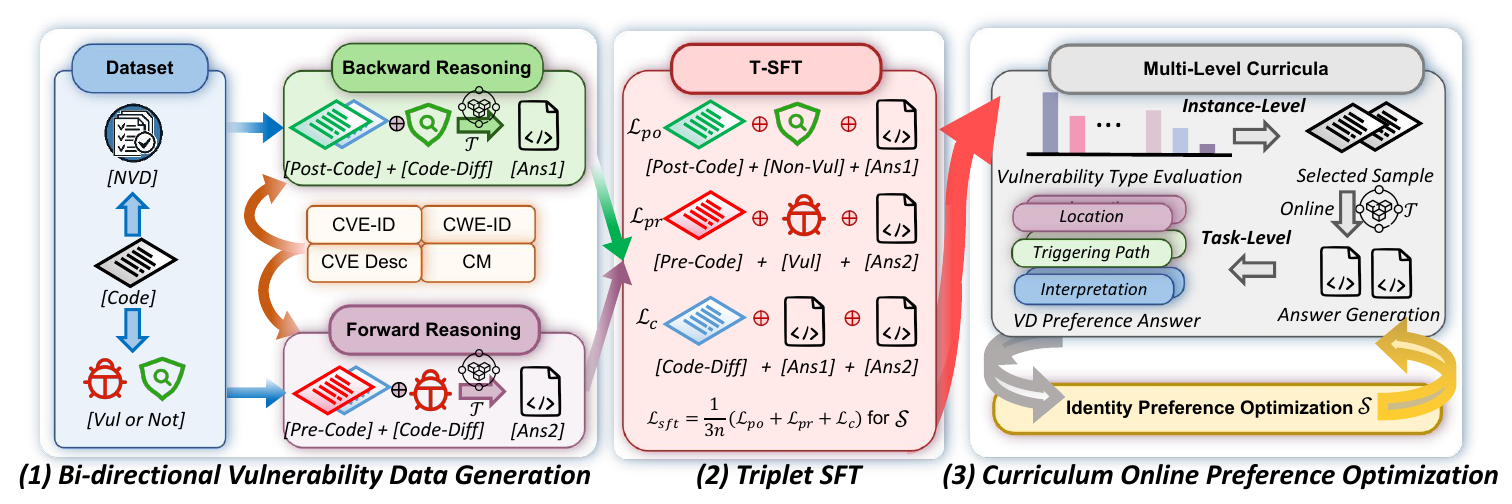}
    \caption{\textbf{The overview of \tool.} \textit{First,} \tool takes a vulnerable code and its corresponding fixed code as inputs to generate backward and forward reasoning answers by model $\mathcal{T}$. \textit{Then}, we train the model $\mathcal{S}$ by the pair of pre-code, post-code, code-diff, and corresponding answers. \textit{Finally}, we undergo multiple rounds of task- and instance-level preference data construction for COPO to optimize model $\mathcal{S}$ continuously.} 

\label{architecture}
\end{figure*}

\paragraph{Software Vulnerability Detection:}
CodePLMs~\cite{DBLP:conf/msr/FuT22/linevul, DBLP:conf/emnlp/FengGTDFGS0LJZ20/codebert, DBLP:journals/tse/ZhangLHXL23/EPVD} and LLMs~\cite{GPT4, ChatGPT} are widely used in vulnerability detection and can be classified into prompt-based and fine-tuning-based methods. The prompt-based method~\cite{DBLP:journals/corr/abs-2310-09810/llm4vul, DBLP:journals/corr/abs-2403-17218, DBLP:journals/corr/abs-2404-02525} has demonstrated effective performance on datasets specific to certain types of vulnerabilities. For instance, 
\citet{DBLP:journals/pacmpl/LiHZQ24/LIFT} propose Lift, which integrates static analysis with LLMs and focuses particularly on the detection of use-before-initialization vulnerabilities in the Linux kernel. Similarly, Sun et al. propose LLM4Vuln~\cite{DBLP:journals/corr/abs-2401-16185/LLM4Vuln} and only target the smart contract vulnerabilities. Additionally, Ding et al.~\cite{DBLP:journals/corr/abs-2403-18624/primevul} explored the use of chain-of-thought prompts, which have shown promising results in enhancing detection capabilities.
The fine-tuning-based~\cite{DBLP:journals/tse/WenGLWLL24, DBLP:conf/msr/FuT22/linevul} approaches add a randomly initialized binary classification head to the language model and jointly optimize all weights based on ground-truth labels. 
With the development of the Transformer architecture, encoder-only Transformers, such as CodeBERT~\cite{DBLP:conf/emnlp/FengGTDFGS0LJZ20/codebert}, and encoder-decoder Transformers, such as UniXcoder~\cite{DBLP:conf/acl/GuoLDW0022/unixcoder}, serve as foundational backbones that are further fine-tuned to enhance their efficacy in identifying software vulnerabilities. Additionally, recent studies have explored the vulnerability capabilities of larger LMs~\cite{DBLP:conf/acl/Du0ZXJLS024/vulllm}, such as code-llama~\cite{DBLP:journals/corr/abs-2308-12950/codellama} and llama~\cite{DBLP:journals/corr/abs-2302-13971/llama}. 
These approaches mainly learn representations from the semantic perspective~\cite{cheng2022bug}. In this work, we propose the first fully automated pipeline to synthesize vulnerability reasoning data.
Furthermore, we develop a novel two-stage framework \tool on our synthetic reasoning data.

\paragraph{Preference Optimization for Code:}
Preference optimization techniques have recently been widely used in enhancing the output quality of LLMs across various natural language-related tasks~\cite{DBLP:conf/icml/XuSCTSDM024/po1, DBLP:conf/recsys/AustinKTS24/po2, DBLP:journals/corr/abs-2410-15595/po3}. These techniques have also been extended to code-related tasks, such as code generation~\cite{DBLP:journals/corr/abs-2410-05605/codedpo}. For instance, Code-Optimize~\cite{DBLP:journals/corr/abs-2406-12502/CodeOptimise} has adopted Direct Preference Optimization (DPO)~\cite{DBLP:conf/nips/RafailovSMMEF23/dpo} as its preferred fine-tuning method, owing to its simplicity and broad acceptance. PPOCoder~\cite{DBLP:journals/tmlr/ShojaeeJTR23/PPOcoder} has introduced a novel framework for code generation that effectively integrates CodePTMs with Proximal Policy Optimization (PPO)~\cite{DBLP:journals/corr/SchulmanWDRK17/ppo}. Additionally, PLUM~\cite{DBLP:conf/acl/ZhangLLLJ23/PLUM} has developed methods to generate tests and utilize Knowledge Transfer Optimization (KTO)~\cite{DBLP:journals/corr/abs-2402-01306/KTO}. However, there have been no studies exploring the application of preference optimization to address vulnerability-related tasks. This gap can be attributed to the inherent need for objective correctness in vulnerability management~\cite{DBLP:journals/isjgp/Nyanchama05}, contrasting with the often subjective nature of preferences in natural language tasks. Moreover, constructing vulnerability data is costly and challenging to procure~\cite{DBLP:journals/corr/abs-2301-05456/dataquality, DBLP:journals/corr/abs-2308-10523/PILOT}, complicating efforts to generate feedback through test cases.
In this paper, 
we design the multi-level curricula and optimization phases to enhance the effectiveness of vulnerability-specific preference optimization.

\section{Proposed Framework}\label{sec:med}

We provide an overview of \tool workflow in Figure~\ref{architecture}. \tool mainly consists of three modules: 
(1) Bi-directional Vulnerability Data (\moduleA) Generation, (2) Triplet Supervised-Fine-Tuning (\moduleB), and (3) Curriculum Online Preference Optimization (\moduleC).
\subsection{BVD Generation}
\label{sec:ivd}
For vulnerability detection,  \tool is designed to enhance the vulnerability detection capabilities in real-world scenarios, where the detector determines whether a code snippet contains a vulnerability and further explains the reason. However, directly deploying LLMs $\mathcal{T}$ for vulnerability detection in real-world scenarios is still challenging due to the lack of high-quality reasoning data related to vulnerabilities in their pre-training corpus.
To bridge this gap, we use the original vulnerability dataset $\mathcal{D}$ with the model $\mathcal{T}$ to generate the vulnerability interpretation data by forward and backward reasoning to produce an augmented dataset $\mathcal{D}_\text{aug}$. 
Specifically, we select a vulnerability patch dataset denoted as $\mathcal{D} = \{(Pr^{(i)}_{c}, Po^{(i)}_{c})\}_{i=1}^n$ of $n$ samples, where each sample comprises a pre-code $Pr^{(i)}_{c}$ (i.e., labeled as vulnerable) and its corresponding post-code $Po^{(i)}_{c}$ (i.e., labeled as non-vulnerable). 

Following the prior work~\cite{DBLP:journals/corr/abs-2411-19865}, we design both forward and backward reasoning processes. The \emph{forward reasoning} process aims to deduce the causes that trigger
the vulnerability in the code. In contrast, the \emph{backward reasoning} process seeks to understand why the code changes in the fixed code prevent the recurrence of the vulnerability and outlines the steps taken to rectify the vulnerability. As illustrated in Figure~\ref{architecture}, the model 
analysis leverages vulnerability information, including code-diff, CVE-ID~\cite{CVE}, CWE-ID~\cite{CWE}, CVE descriptions, and commit messages from the NVD~\cite{nvd}. These descriptions provide
relevant background and context
of the vulnerabilities and their fixes, which are
beneficial for understanding the root causes. Specifically, \tool takes a code, corresponding code change in the patch, and vulnerability information as inputs to generate forward and backward reasoning traces.
The detailed example is shown in Appendix~\ref{appendix:answer}. 
We have the augmented vulnerability reasoning dataset $\mathcal{D}_\text{aug}$, where each data in $\mathcal{D}_\text{aug}$ consists of 
$(Pr^{(i)}_{c}, Po^{(i)}_{c}, Pr^{(i)}_{a}, Po^{(i)}_{a}, Di^{(i)}_{c}, Di^{(i)}_{a})$. It denotes the pre-code, forward reasoning answer for pre-code, post-code, backward reasoning answer for post-code, code-diff, and the corresponding answer, respectively. 

\subsection{Triplet Supervised Fine-Tuning}
\label{sec:tsft}

In order to enhance the LLM's ability to capture vulnerability patterns,
we analyze the relationship among pre-code, post-code, and code-diff and construct a triplet loss to fine-tune the model $\mathcal{S}$ with the previously augmented dataset $\mathcal{D}_\text{aug}$ as below:\looseness=-1
\begin{align}\label{eq:loss}
&\mathcal{L}_{\text{T-SFT}} = \mathbb{E}_{\mathcal{D}_\text{aug}} \Bigl[ 
\underbrace{\ell(\mathcal{S}(Q,Pr^{(i)}_{c}), \smash{Pr^{(i)}_{a}})}_{\text{(i) Pre-code}}+ \\
&
\underbrace{\ell(\mathcal{S}(Q,Po^{(i)}_{c}), \smash{Po^{(i)}_{a}})}_{\text{(ii) Post-code}}
+\underbrace{\ell(\mathcal{S}(Q,Di^{(i)}_{c}), \smash{Di^{(i)}_{a}})}_{\text{(iii) Code-Diff}}
\Bigr] \nonumber
\end{align}
where $Q$ denotes the question for vulnerabiliy detection and $\ell$ is the cross-entropy loss.
Specifically, the loss $\mathcal{L}_{\text{T-SFT}}$ is composed of three losses that make full use of our augmented data: 
(i) the pre-code part for learning vulnerability patterns, (ii) the post-code part for learning how to fix the vulnerability, and (iii) the code-diff part for focusing on vulnerability-specific code changes rather than the similar code semantics.
We aim to let the model concentrate on the vulnerability patterns in a multi-task learning approach. Different components in Eq.~\ref{eq:loss} are treated equally because understanding how vulnerabilities are triggered and how to fix them is equally important for vulnerability detection.

\begin{algorithm}[t!]
\caption{Training Procedure of COPO [\ref{sec:copo}]}
\label{alg:COPO}
	\begin{algorithmic}[1]
    	\State \textbf{initialize:~}$r=0$, $\mathcal{D}^{(0)}\leftarrow\varnothing$, Model $\pi \leftarrow s$, Regularize-Parameter $\tau\in R^+$, Type List $l$
	    \State \textbf{input:~} Round $\mathcal{C}$, Generation Model $\mathcal{T}$,  Eval Set $E$, Training Set $\mathcal{E}$, Reference Policy $\pi_{\text{ref}}$,
            
            \While{$r<C$} 
                \State Test $E$ via $s$  \Comment{\textcolor{mydarkred}{Instance-level curriculum}}
                \State Get $acc_{r}=[t_{1}^{r},t_{2}^{r}\dots,t_{n}^{r}$] via $l$
                \State $probabilities \gets [1 - t_{i}^{r} \text{ for } t_{i}^{r} \text{ in } acc_{r}]$
\State $S_s^{(r)} \gets \{\}$
\For{$data \in \mathcal{D}$}
    \State $type_d \gets \text{Get\_vul\_type}(data, l)$ 
    \State $p_d \gets probabilities[type_d]$
    \State $r_d \gets \text{random between 0 and 1}$
    \If{$r_d < p_d$}
        \State $S_s^{(r)} \gets S_s^{(r-1)} \cup \{data\}$
    \EndIf
\EndFor
                \State $\mathcal{D}^{(r)} = (y_w, y_l, \bigcup_{1}^{r} x_r)) \leftarrow S_s^{(r)}$
                
                \State Decomposition \Comment{\textcolor{mydarkred}{Task-level curriculum}} 
                \State Generate the preferred and dispreferred generations $y_w$ and $y_l$ via $\mathcal{T}$ and 
                \State Optimization Process by Eq.~(\ref{eq:ipo1}) and (\ref{eq:ipo2}) with $y_w$, $y_l$, $\pi$, $\pi_{\text{ref}}$ and $\tau$
                \Comment{\textcolor{mydarkred}{Optimization}} 
                \State $r\leftarrow r+1$
            \EndWhile
            \State \textbf{output:~}$\pi_{\theta^{*}}$
	\end{algorithmic}
    \label{alg1}
\end{algorithm}

\subsection{Curriculum Online Preference Optimization (COPO)}\label{sec:copo}
To correctly distinguish vulnerabilities with fixed code,
we develop the \moduleC module that is structured into two distinct phases: \textbf{Curriculum} and \textbf{Optimization}, as elaborated in our Algorithm~\ref{alg:COPO}.

\textbf{Multi-level Curricula. } 
It is evident that instances of different vulnerability types
are unevenly distributed, as shown in Appendix~\ref{appendix:distribution}.
The imbalanced distribution can lead to suboptimal performance, which stimulates the development of a curriculum-based method focusing more on types of vulnerabilities.
According to the features of vulnerability detection, we build instance-level and task-level curricula.
\textit{(1) Instance-level curriculum.} We first construct a list of vulnerability types based on the CWE~\cite{CWE}, denoted as $L$ and divide an evaluation set $E$. Detailed information is shown in Table~\ref{appendix:distribution}. 
Based on the current accuracy of the model $S$ (as mentioned in Sec.~\ref{sec:tsft}) in identifying different vulnerabilities within $E$, we filter instances in Lines 4-6. The lower the accuracy for the vulnerability type, the more likely it is that instances of this type from dataset $D$ will be selected as preference instances $data$ (Lines 6-13). The instance $data$ for the current round is then merged with the instances from the previous iteration $S_s^{(r-1)}$, to form a new preference dataset $S_s^{(r)}$.
\textit{(2) Task-level curriculum. } Then, 
to mitigate the scarcity of vulnerability data, we further generate a 3x larger dataset by breaking down the established one. In particular, based on three easy-to-hard vulnerability comprehension tasks,
we decompose each preference instance into three different tasks: location of the vulnerable line, trigger path of program analysis, and root cause interpretation. We also use the \moduleA to regenerate instances by model $\mathcal{T}$ (as mentioned in Section~\ref{sec:ivd}). 
For each pre-code, we denote the explanation data as the referred responses $y_w$ and
the corresponding fixed code as dispreferred responses $y_i$.  It is important to note that each instance is exclusively utilized for one task in any iteration round, although the same instance may be employed across different rounds and for varied task assignments.

\textbf{Optimization. } Finally, we improve Identity Preference Optimization (IPO)~\cite{DBLP:conf/aistats/AzarGPMRVC24/ipo}.
It avoids model $\mathcal{S}$ to overfit the preference data 
in the limited vulnerability dataset.
Based on the IPO, we iteratively optimize the curriculum criteria each round which is calculated as follows:
\begin{equation}
\label{eq:ipo1}
h_\pi(y_w, y_l,x ) = \log\left( \frac{\pi(y_w|\bigcup_{1}^{r} x_r)\piref(y_l|\bigcup_{1}^{r} x_r)}{\pi(y_l|\bigcup_{1}^{r} x_r)\piref(y_w|\bigcup_{1}^{r} x_r)} \right)
\end{equation}
\begin{equation}
\label{eq:ipo2}
\underset{(y_w,y_l,x)\sim D}{\mathbb E} \left(h_\pi(y_{w}, y_{l},x) - \frac{\tau^{-1}}2\right)^2 .
\end{equation}
Specifically, \moduleC iteratively
optimizes the model to align with preferences data, which regress the gap between log-likelihood ratios $\log(\pi(y_{w})/\pi(y_{l}))$ and $\log(\piref(y_{w})/\piref(y_{l}))$  to regularization parameter $\frac{\tau^{-1}}2$. 


\begin{table*}[t]
\centering
\setlength{\tabcolsep}{2mm}
\renewcommand{\arraystretch}{0.95}
\caption{
\textbf{Evaluation results of \tool compared with vulnerability detection baselines on the PrimeVul and SVEN datasets.} 
COT = Chain of Thought. SFT = Supervised Fine-Tuning. The prompt template used in our experiments follows the approach outlined by Ding et al.~\cite{DBLP:journals/corr/abs-2403-18624/primevul}.
The \textbf{highest score} for each metric in the same dataset are highlighted in \textbf{bold} text. The  {\color[HTML]{FE0000} {($\uparrow$)}} / {\color[HTML]{009901} {($\downarrow$)}}  represents the performance of the \tool compared with the best-performing method on this metric. 
\textbf{\tool significantly surpasses all SOTA baselines in the PrimeVul and SVEN datasets. When integrated with Qwen2.5-Coder-7B-Instruct, \tool achieves the highest performance.}
}
\resizebox{0.94\textwidth}{!}{
\begin{tabular}{lc|ccc|ccc}
\toprule
\rowcolor[HTML]{DEDEDE}
  & \textbf{Dataset} & \multicolumn{3}{c|}{\textbf{PrimeVul}~\cite{DBLP:journals/corr/abs-2403-18624/primevul}}  & \multicolumn{3}{c}{\textbf{SVEN}~\cite{DBLP:conf/ccs/HeV23/SVEN}}      \\
\rowcolor[HTML]{DEDEDE}
   \multirow{-2}{*}{\textbf{Method}}                        & \multicolumn{1}{c|}{\textbf{Type}}   & \textbf{Accuracy $\uparrow$} & \textbf{F1 Score $\uparrow$}   & \textbf{VP-S $\uparrow$}  & \textbf{Accuracy $\uparrow$} & \textbf{F1 Score $\uparrow$}   & \textbf{VP-S $\uparrow$}  \\
   \toprule
CodeBERT~\cite{DBLP:conf/emnlp/FengGTDFGS0LJZ20/codebert}                  & SFT     & 51.03    & 16.47      & 2.07  & 51.77    & 49.79      & 3.53  \\
UniXCoder~\cite{DBLP:conf/acl/GuoLDW0022/unixcoder}                 & SFT     & 49.89    & 16.48      & -0.23 & 51.90    & 43.45      & 3.80  \\
LineVul~\cite{DBLP:conf/msr/FuT22/linevul}                   & SFT     & 49.77    & 17.70      & -0.46 & 51.90    & 40.40      & 3.80  \\
Llama3.1-70B-Instruct~\cite{DBLP:journals/corr/abs-2407-21783/llama3.1}              & COT     & 49.77    & 55.90      & -0.46 & 50.00    & 31.34      & 0.00  \\
Qwen2.5-32b-Coder-Instruct~\cite{DBLP:journals/corr/abs-2409-12186/qwen2.5}                  & COT     & 50.00    & 26.64      & 0.00  & 51.36    & 29.25      & 2.72  \\
GPT-4~\cite{GPT4}                   & COT     & 51.72    & 43.40      & 3.44  & 51.36    & 56.76      & 2.72  \\
\midrule
 \multirow{2}{*}{Qwen2.5-Coder-7B-Instruct~\cite{DBLP:journals/corr/abs-2409-12186/qwen2.5}}  & COT     & 49.77    & 29.86      & -0.46 & 52.31    & 27.63      & 4.62  \\
                          &    \textbf{\tool}      & \textbf{58.05} \color[HTML]{FE0000} {($\uparrow$6.33\%)}   & 63.83\color[HTML]{FE0000} {($\uparrow$7.93\%)}     & \textbf{16.09} \color[HTML]{FE0000} {($\uparrow$12.65\%)} & \textbf{63.72}  \color[HTML]{FE0000} {($\uparrow$11.82\%)}  & \textbf{69.49}  \color[HTML]{FE0000} {($\uparrow$12.73\%)}    & \textbf{27.4}4 \color[HTML]{FE0000} {($\uparrow$23.64\%)}\\
                        \midrule
 \multirow{2}{*}{Llama-3.1-8B-Instruct~\cite{DBLP:journals/corr/abs-2407-21783/llama3.1} }     & COT     & 48.16    & 40.58      & -3.68 & 50.00    & 46.82      & 0.00  \\
                        &    \tool      & 56.21 \color[HTML]{FE0000} {($\uparrow$4.49\%)}   & 54.91  \color[HTML]{009901} {($\downarrow$-0.99\%)}    & 12.41\color[HTML]{FE0000} {($\uparrow$8.97\%)} & 61.82 \color[HTML]{FE0000} {($\uparrow$9.92\%)} &	60.70 \color[HTML]{FE0000} {($\uparrow$3.94\%)} &	23.65 \color[HTML]{FE0000} {($\uparrow$19.85\%)} \\
                          \midrule
 \multirow{2}{*}{StarCoder2-7B~\cite{DBLP:journals/corr/abs-2402-19173/starcoder2} }            & COT     & 50.11    & 0.91       & 0.23  & 50.00    & 0.00       & 0.00  \\
                          &    \tool      & 55.63 \color[HTML]{FE0000} {($\uparrow$3.91\%)}   & \textbf{68.74}\color[HTML]{FE0000} {($\uparrow$12.84\%)}      & 11.26\color[HTML]{FE0000} {($\uparrow$7.82\%)} & 57.34\color[HTML]{FE0000} {($\uparrow$5.44\%)}    & 53.96\color[HTML]{009901} {($\downarrow$2.80\%)}      & 14.68\color[HTML]{FE0000} {($\uparrow$10.88\%)} \\
\bottomrule
\label{RQ1}
\end{tabular}}

\end{table*}

\section{Experimental Setup}\label{sec:setup}

In vulnerability detection, the existing methods focus on C/C++ due to their widespread use in real-world programming and the presence of well-known vulnerabilities that vulnerability researchers correctly label. We choose two widely-used and higher-quality vulnerability datasets, including PrimeVul~\cite{DBLP:journals/corr/abs-2403-18624/primevul}, and SVEN~\cite{DBLP:conf/ccs/HeV23/SVEN}, to provide a more accurate evaluation under real-world conditions.
PrimeVul employs a data labeling technique and contains 140 CWEs and 6,968 vulnerable samples across 755 projects and 6,827 commits. 
We classified the vulnerability types based on the CWE~\cite{CWE} and GraphSPD~\cite{DBLP:conf/sp/WangWSJWL23/GraphSPD}. 
As shown in Appendix~\ref{cwe_dataset}, the ratio distribution is imbalanced and contains different probabilities for instance-level curriculum.
SVEN dataset manually vets vulnerabilities from multiple repositories, containing 416 vulnerable and fixed codes from real-world C/C++ projects, with the highest 94\% reported accuracy. 

\subsection{Metrics}
Following the previous work~\cite{DBLP:conf/kbse/WenWGWLG23, DBLP:conf/msr/FuT22/linevul}, we choose the Accuracy and F1 to evaluate \tool's performance. Because the F1 may bias towards models which predict vulnerable more often~\cite{DBLP:journals/corr/abs-2404-02525}, we
propose the Vulnerability Pair-Score (VP-S$=\frac{Correct_{pair}-Wrong_{pair}}{All_{pair}}$) to evaluate the performance for vulnerability detection. 
$Correct_{pair}$ and $Wrong_{pair}$ denote the correctly and inversely predict the ground-truth labels, respectively.

\begin{table*}[t]
\centering
\setlength{\tabcolsep}{3.5mm}
\renewcommand{\arraystretch}{0.98}
\caption{
\textbf{Ablation study.} The experimental results of \tool and corresponding variants in PrimeVul and SVEN datasets. 
The ``\textit{w/o \moduleA + \moduleB}'' uses the original data and the full supervised fine-tuning for training.
The ``\textit{w/o \moduleC}'' replaces the \moduleC (described in Algorithm~\ref{alg:COPO}).
Please refer to Section.~\ref{sec:ablation} for a thorough discussion.}
\resizebox{0.92\textwidth}{!}{
\begin{tabular}{lc|ccc|ccc}
\toprule
\rowcolor[HTML]{DEDEDE}
  & \textbf{Dataset} & \multicolumn{3}{c|}{\textbf{PrimeVul}}  & \multicolumn{3}{c}{\textbf{SVEN}}      \\
\rowcolor[HTML]{DEDEDE}
   \multirow{-2}{*}{\textbf{Base Model}}                        & \multicolumn{1}{c|}{\textbf{Varient}}   & \textbf{Accuracy $\uparrow$} & \textbf{F1 Score $\uparrow$}   & \textbf{VP-S $\uparrow$}  & \textbf{Accuracy $\uparrow$} & \textbf{F1 Score $\uparrow$}   & \textbf{VP-S $\uparrow$}  \\
   \toprule
GPT-4~\cite{GPT4}                    & -    & 51.72    & 43.40      & 3.44  & 51.36    & 56.76      & 2.72  \\
\midrule
\multirow{3}{*}{Qwen2.5-Coder-7B-Instruct~\cite{DBLP:journals/corr/abs-2409-12186/qwen2.5}} & w/o \moduleA + \moduleB      & 51.03    & 28.28      & 2.07  & 50.00    & 0.00       & 0.00  \\
                                           & w/o \moduleC    & 55.75    & 56.50      & 11.49 & 54.21    & 61.83      & 8.43  \\
                                           & \tool & \textbf{58.05}    & \textbf{63.83}      & \textbf{16.09} & \textbf{63.72}    & \textbf{69.49}      & \textbf{27.44} \\
                                           \midrule
\multirow{3}{*}{Llama-3.1-8B-Instruct~\cite{DBLP:journals/corr/abs-2407-21783/llama3.1}}     & w/o \moduleA + \moduleB      & 50.00    & 0.00       & 0.00  & 50.00    & 0.00       & 0.00  \\
                                           & w/o \moduleC    & 55.16    & 59.79      & 10.34 & 60.87    & 63.82      & 21.74 \\
                                           & \tool & \textbf{56.21}    & \textbf{54.91}      & \textbf{12.41} & \textbf{61.82}    & \textbf{60.70}      & \textbf{23.65} \\
                                           \midrule
\multirow{3}{*}{StarCoder2-7B~\cite{DBLP:journals/corr/abs-2402-19173/starcoder2}}             & w/o \moduleA + \moduleB      & 50.00    & 0.00       & 0.00  & 50.00    & 0.00       & 0.00  \\
                                           & w/o \moduleC    & 52.87    & 58.59      & 5.74  & 54.21    & 61.83      & 8.43  \\
                                           & \tool & \textbf{55.63}    & \textbf{68.74}      & \textbf{11.26} & \textbf{57.34}    & \textbf{53.96}      & \textbf{14.68} \\
\bottomrule
\label{RQ2}
\end{tabular}}
\end{table*}

\subsection{Training and Inference Details}
For the training of \moduleB, we employ the full SFT and train each LLM for three epochs. In contrast, for \moduleC, we utilize the LoRA~\cite{DBLP:journals/corr/abs-2406-03136/lora}, which updates only a subset of the parameters in the base model while keeping the remaining components unchanged. We conduct three rounds of preference optimization, and each round is trained for one epoch. The maximum sequence length is set to 2048 tokens. Further details about the
hyperparameters used in our training procedure
are available in Table~\ref{appendix:hyper} of the Appendix~\ref{appendix:hyperparameters}.


\section{Experimental Results}\label{sec:result}

\subsection{Comparison with State-of-the-Art}
In this section, we compare \tool with nine other representative vulnerability detection methods on the PrimeVul and SVEN datasets, spanning three widely-used vulnerability detection methods, three larger LLMs (exceeding 30B), and three base models. Detailed in Appendix~\ref{appendix:baseline}.

\textbf{(1) \tool sets new SOTA performance in software vulnerability detection.} 
The experimental results presented in Table~\ref{RQ1} demonstrate that \tool consistently outperforms all baseline methods across all datasets. Specifically, when integrated into the Qwen2.5-Coder-7B-Instruct, \tool exhibits superior performance across all metrics. It achieves an accuracy of up to 58.05\%, an F1 score of 63.83\%, and a VP-Score of 16.09\% on the PrimeVul dataset. This achievement underscores the effectiveness of \tool in capturing vulnerability patterns across complex real-world scenarios, encompassing 140 types of vulnerabilities. Furthermore, \tool exhibits a higher accuracy of 63.72\%, an F1 score of 69.49\%, and a VP-Score of 27.44\% on the SVEN dataset. This indicates that \tool is equally robust on other datasets that do not require training. Notably, the SVEN dataset includes only nine types of vulnerabilities, which may contribute to the enhanced performance of \tool.

\textbf{(2) \tool is versatile across a variety of models.} We select two types of LLMs, including the CodeLLMs (i.e., Qwen2.5 and StarCoder2) and the general LLM (i.e., LLama-3.1), as base models to validate the extensive applicability of \tool.
As illustrated in Table~\ref{RQ1}, \tool's performance remains significantly superior to previous methods in most cases, regardless of the base model. This can be attributed to \tool's guide model to capture vulnerability patterns in code changes, focusing on the triggering reasoning of vulnerabilities rather than merely their semantics.
On the other hand, some vulnerabilities that are not typically well-handled by LLMs are directly addressed by COPO, which ensures performance enhancements. This finding highlights the potential advantages of \tool in real-world scenarios, where the types and samples of vulnerabilities are continuously expanding.

\definecolor{mypink}{RGB}{251, 126, 123}
\definecolor{mypurple}{RGB}{148, 103, 189}
\definecolor{myorange}{RGB}{255, 180, 105}
\definecolor{mygreen}{RGB}{78, 194, 194}
\begin{figure*}[t]
\centering
	\includegraphics[width=0.95\textwidth]{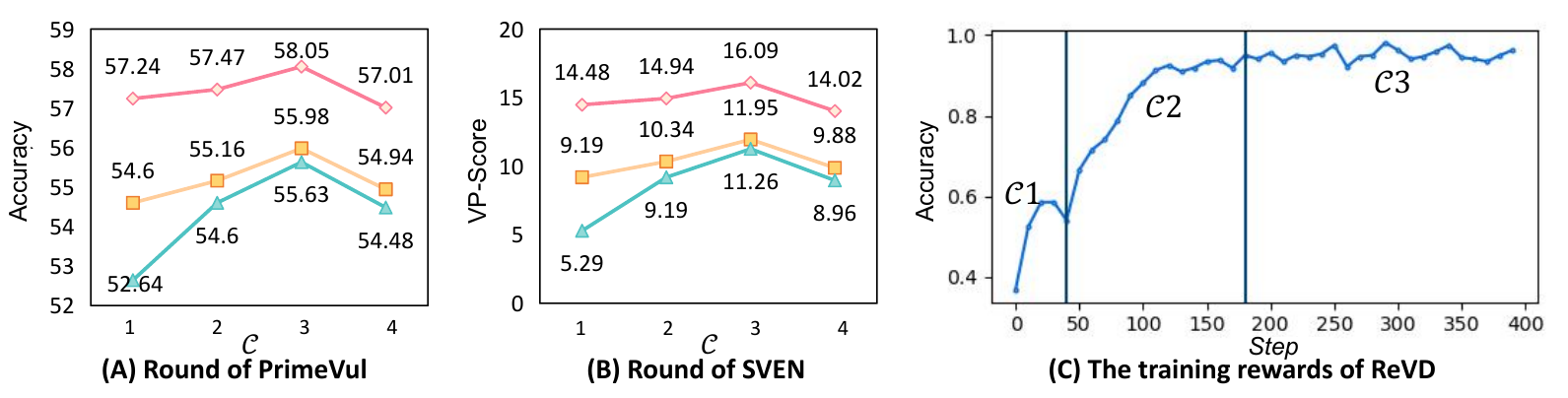}
    \caption{\textbf{The impact of different rounds of \moduleC.} \textit{Left}: The performance on PrimeVul.  \textit{Middle}: The performance on SVEN. \textit{Right}: The training reward accuracies of \tool across \moduleC steps. The \textcolor{mypink}{pink},  \textcolor{myorange}{orange}, and \textcolor{mygreen}{green} lines denote the \tool involved with \textcolor{mypink}{Qwen2.5-Coder-7B-Instruct}, \textcolor{myorange}{Llama-3.1-8B-Instruct} and \textcolor{mygreen}{StarCoder2-7B} metrics, respectively. $C$ denotes the number of \textbf{C}urriculum Round. }
\label{RQ3}
\end{figure*}
\subsection{Ablation Study}
\label{sec:ablation}
In this section, we explore the impact of different modules of \tool across two datasets, including the \moduleA, \moduleB, and \moduleC.
The experimental results are shown in Table~\ref{RQ2}.

\textbf{(1) Bi-directional vulnerability reasoning
data improves the reasoning abilities of LLMs for software vulnerability detection.} 
To assess the importance of the bi-directional vulnerability interpretation data, we deploy one variant (i.e., w/o \moduleA + \moduleB) by using the original data proposed by PrimeVul. Due to the \moduleB being designed for \moduleA, this variant only uses the full supervised fine-tuning for training. 
The results show that \tool with the bi-directional vulnerability interpretation data significantly outperforms its counterpart. It averagely improves 6.29\% of PrimeVul, and 10.96 \% of SVEN in accuracy.
This finding is crucial as it indicates that existing LLMs can reason about vulnerabilities, but rather lack the domain-specific knowledge required to effectively invoke such reasoning. \moduleA and \moduleB are fundamental components of \tool, providing accurate interpretations of vulnerabilities and contributing substantially to its enhanced performance. 

\textbf{(2) \moduleC enables more effective vulnerability detection.}
To evaluate the effectiveness of \moduleC in Algorithm~\ref{alg1}, we also conduct a variant without the \moduleC (i.e., w/o \moduleC).
The results reveal that the variant average exhibits a decrease of 2.04\% on Primevul and 4.53\% on SVEN in accuracy, suggesting that \moduleC contributes to enhancing the upper performance bound of \tool.
This improvement can be attributed to \moduleC's ability to effectively identify the samples and vulnerability types that the current model struggles with. It then strategically selects these samples in subsequent training phases, thereby optimizing the performance of \tool.

\subsection{\moduleC Round}

Considering the limited labeled data of vulnerability and training efficiency, we conduct experiments with the number of training rounds in \moduleC, which are proportional to the training time. As shown in Figure~\ref{RQ3} (A) and (B), we observe that the \tool performance increases as the round increases in the initial phase, reaching its peak at round 3. 
However, further increases in training rounds result in a decline in performance. It can be attributed to the limited pool of training samples, despite our efforts to augment the data through task answer decomposition and setting the temperature to 1 for generating interpretative answers online. In addition, we also find that the reward accuracies have
Furthermore, we observe that the reward accuracies exhibit oscillations in round 3 in Figure~\ref{RQ3} (C). The continuation of curriculum training has a minimal impact on this situation.

\definecolor{mypink}{RGB}{251, 126, 123}
\definecolor{myorange}{RGB}{255, 180, 105}
\definecolor{myyellow}{RGB}{255, 205, 155}
\definecolor{mygreen}{RGB}{78, 194, 194}



\begin{figure}[t]
\centering
	\includegraphics[width=0.48\textwidth]{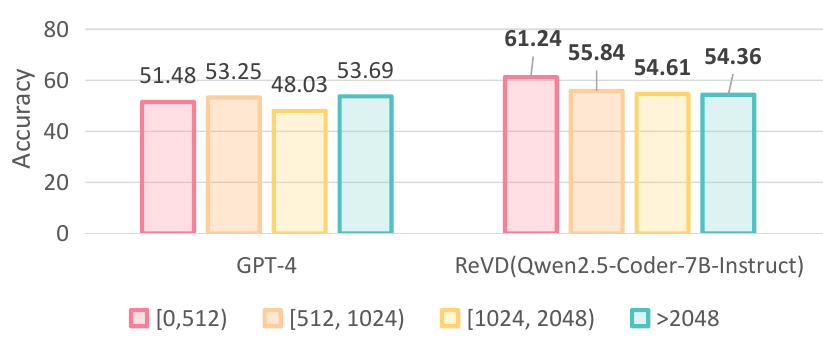}
    \caption{\textbf{The accuracy of \tool under varying token length.} The \textcolor{mypink}{pink},  \textcolor{myorange}{orange}, \textcolor{myyellow}{yellow}, and \textcolor{mygreen}{green} denote the performance for token numbers containing \textcolor{mypink}{less than 512 tokens}, \textcolor{myorange}{between 512 and 1024 tokens}, \textcolor{myyellow}{between 1024 and 2048 tokens}, and \textcolor{mygreen}{more than 2048 tokens}.
   } 
\label{RQ4}
\end{figure}
\subsection{Token Length}

The capability of LLMs to accurately detect vulnerabilities across varying code lengths is essential for addressing real-world challenges. To evaluate this, we conduct a supplementary experiment on PrimeVul, where we classify the token lengths of examples into four categories: less than 512 tokens (ratio of 38.85\%), 512-1024 tokens ((ratio of 26.55\%), 1024-2048 tokens (ratio of 17.47\%), and more than 2048 tokens (ratio of 17.13\%).

We compare the performance of \tool with the best-performing LLM, GPT-4, and the results are depicted in Figure~\ref{RQ4}. We observe that \tool exhibits a slight decline in performance as the number of tokens increases, yet it significantly outperforms GPT-4 across all token ranges. Additionally, we find the following observations:
(1) \tool, integrated with Qwen-2.5-7B-Instruct, generally performs best in most scenarios (in 3 out of 4 cases).
(2) The capability for vulnerability detection remains relatively consistent across the 512-2048 token range, with a notable decline only occurring beyond 2048 tokens, indicating a need for more data involving longer token lengths. The complete results of \tool are provided in Appendix~\ref{appendix:rq4}.

\subsection{Curriculum Strategy}
To evaluate the efficacy of instance- and task-level curriculum strategies, we conducted two variants.
The ``\textit{w/o IC}'' variant uses the equal selection criteria for all samples (i.e., $p_d=0.5$) without considering the vulnerability types.
The ``\textit{w/o TC}'' variant only uses one task for preference optimization.

As shown in Table~\ref{RQ5}, the results indicate that the variants do not achieve the same performance as the \tool. This suggests that the instance- and task-level curriculum strategy enhances the performance of models.
Additionally, we observe that the impact of the curriculum strategy on Qwen2.5 is minimal, which may be attributed to the inherently superior capabilities of the base model.

\begin{table}[t]
\centering
\setlength{\tabcolsep}{3mm}
\renewcommand{\arraystretch}{0.95}
\caption{
\textbf{The effectiveness of the instance-level and task-level curriculum.} 
}
\resizebox{0.49\textwidth}{!}{
\begin{tabular}{lc|ccc}
\toprule
\rowcolor[HTML]{DEDEDE}
  & \textbf{Dataset} & \multicolumn{3}{c}{\textbf{PrimeVul}}        \\
\rowcolor[HTML]{DEDEDE}
   \multirow{-2}{*}{\textbf{Base Model}}                        & \multicolumn{1}{c|}{\textbf{Varient}}   & \textbf{Accuracy $\uparrow$} & \textbf{F1 Score $\uparrow$}   & \textbf{VP-S $\uparrow$}   \\
   \toprule
\multirow{3}{*}{Qwen2.5-7B-Coder-Instruct} & w/o IC   & 57.47    & 61.70      & 14.94 \\
& w/o TC & 57.13    & 60.36    & 14.25  \\
                                           & \moduleC & 58.05    & 63.83      & 16.09 \\
                                    \midrule
\multirow{3}{*}{Llama-3.1-8B-Instruct}     & w/o IC   & 55.52    & 55.11      & 11.03 \\
& w/o TC & 52.41    & 27.62    & 4.83   \\
                                           & \moduleC & 56.21    & 54.91      & 12.41 \\
                                            \midrule
\multirow{3}{*}{StarCoder2-7B}             & w/o IC   & 53.10    & 53.54      & 6.21  \\
& w/o TC & 51.26    & 23.74    & 2.53   \\
                                           & \moduleC & 55.63    & 68.74      & 11.26 \\
\bottomrule
\label{RQ5}
\end{tabular}}
\end{table}

\section{Conclusion}\label{sec:conclusion}
This paper introduces a novel framework \tool to boost the vulnerability detection of LLMs through vulnerability-specific preference optimization with reasoning-centric data synthesis. 
Specifically, we 
propose the first fully automated pipeline to synthesize vulnerability reasoning data.
Then, we develop a two-stage framework \tool on our synthetic reasoning data, which turns open-source LLMs into strong vulnerability detectors by aligning their coding knowledge with vulnerability semantics, patterns, and types.
Experimental results underscore the effectiveness of \tool for vulnerability detection compared with the state-of-the-art approaches and shed 
novel insights on training more generalizable and versatile LLM experts for vulnerability detection in real-world scenarios.


\section{Limitations}
Due to computing resource constraints, \tool only uses LLMs with sizes of 7B and 8B. This limitation could slightly bias the final performance of the \tool. Additionally, since all vulnerability reasoning samples are less than 4,096 tokens, \tool may struggle to assess vulnerabilities that exceed this length. It may lead to a potential limitation when evaluating longer code snippets for software vulnerability detection. Conducting experiments with a larger computing budget will be our future work. \looseness-1



\bibliography{custom}

\newpage
\appendix

\label{sec:appendix}

\section{Baselines}
\label{appendix:baseline}
\textbf{Pretrained Code Models} We select three prominent pre-trained models:
CodeBERT~\cite{DBLP:conf/emnlp/FengGTDFGS0LJZ20/codebert}, UniXcoder~\cite{DBLP:conf/acl/GuoLDW0022/unixcoder}
and LineVul~\cite{DBLP:conf/msr/FuT22/linevul}. These models use source code as input and are further fine-tuned for the downstream. It is applied to vulnerability detection by fine-tuning.


\textbf{Large Language Models}
We choose two larger open-source LLMs: Llama3.1-70B-Instruct ~\cite{DBLP:journals/corr/abs-2407-21783/llama3.1} and Qwen2.5-32B-Coder-Instruct~\cite{DBLP:journals/corr/abs-2409-12186/qwen2.5} for their proficiency in text and code generation, respectively. Additionally, we also incorporate the closed-source LLMs: GPT-4~\cite{GPT4} for vulnerability detection, given their robust capabilities in handling code-related tasks.

\section{Vulnerability Types Distribution in PrimeVul}
\label{appendix:distribution}
\begin{figure}[h]
	\includegraphics[width=0.45\textwidth]{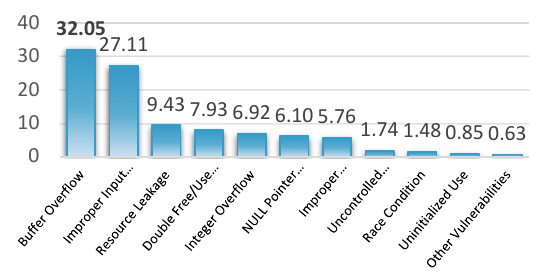}
    \caption{\textbf{Primevul~\cite{DBLP:journals/corr/abs-2403-18624/primevul}'s distribution of vulnerability types.} These vulnerability types are provided by CWE~\cite{CWE} and GraphSPD~\cite{DBLP:conf/sp/WangWSJWL23/GraphSPD} for \moduleC.} 
\label{cwe_dataset}
\end{figure}

In this section, we detail the vulnerability types distribution in PrimeVul~\cite{DBLP:journals/corr/abs-2403-18624/primevul}. They are provided by CWE~\cite{CWE} and GraphSPD~\cite{DBLP:conf/sp/WangWSJWL23/GraphSPD} for \moduleC in Figure~\ref{cwe_dataset}.

\section{Training Details}
\label{appendix:hyperparameters}
In this section, we detail the training details and hyperparameters employed for training Triplet SFT (\moduleB) and curriculum online preference optimization (\moduleC), as outlined in Table~\ref{tab:hyperparameters}. These hyperparameters are primarily adapted for the base model, specifically including Qwen2.5-Coder-7B-Instruct~\cite{DBLP:journals/corr/abs-2409-12186/qwen2.5}, Llama-3.1-8B-Instruct~\cite{DBLP:journals/corr/abs-2407-21783/llama3.1}, and StarCoder2-7B~\cite{DBLP:journals/corr/abs-2402-19173/starcoder2}. We all use Qwen2.5-32B-Coder-Instruct~\cite{DBLP:journals/corr/abs-2409-12186/qwen2.5} as the data generation model $\tau$ for vulnerability detection interpretation and preference data construction.


Our implementation relies extensively on the tools and protocols provided by LLaMA-Factory~\cite{zheng2024llamafactory}. All processes adhere to the standard procedures established by LLaMA-Factory, ensuring consistency and reliability in our training approach.

For inference, to ensure that the experiment is replicable, we employ greedy decoding with \texttt{temperature} = 0 to ensure that results can be reproduced. All experiments are conducted using 8 A100 (40 GB) GPUs.

\begin{table}[]
\caption{Hyperparameters of \tool}
\label{tab:hyperparameters}
\begin{center}
\begin{small}
\begin{tabular}{lcc}
\toprule
\textbf{Hyperparameter} & & \textbf{Value}  \\
\midrule
\multicolumn{3}{c}{\textbf{Triplet SFT}} \\
Max text length   && $2048$ \\
Fine-tuning types && Full \\
Fine-tuning epochs && $3$ \\
Warmup ratio    && $0.1$ \\
Learning rate     && $10^{-5}$ \\
Batch size        && $8$ \\
\midrule

\midrule
\multicolumn{3}{c}{\textbf{\moduleC}} \\

Finetuning types && Lora \\
Pref $\beta$ && $0.1$ \\
Pref Loss&& IPO~\cite{DBLP:conf/aistats/AzarGPMRVC24/ipo} \\
Round && $3$ \\
Learning rate      && $5\times 10^{-6}$ \\
Warmup steps      && $300$ \\
Batch size         && $8$ \\
Training epochs per round && $1$ \\

\bottomrule
\end{tabular}
\end{small}
\end{center}
\label{appendix:hyper}
\end{table}

\section{Complete Experiment Results in Token Length}
\label{appendix:rq4}
\begin{figure*}[t]
\centering
	\includegraphics[width=0.99\textwidth]{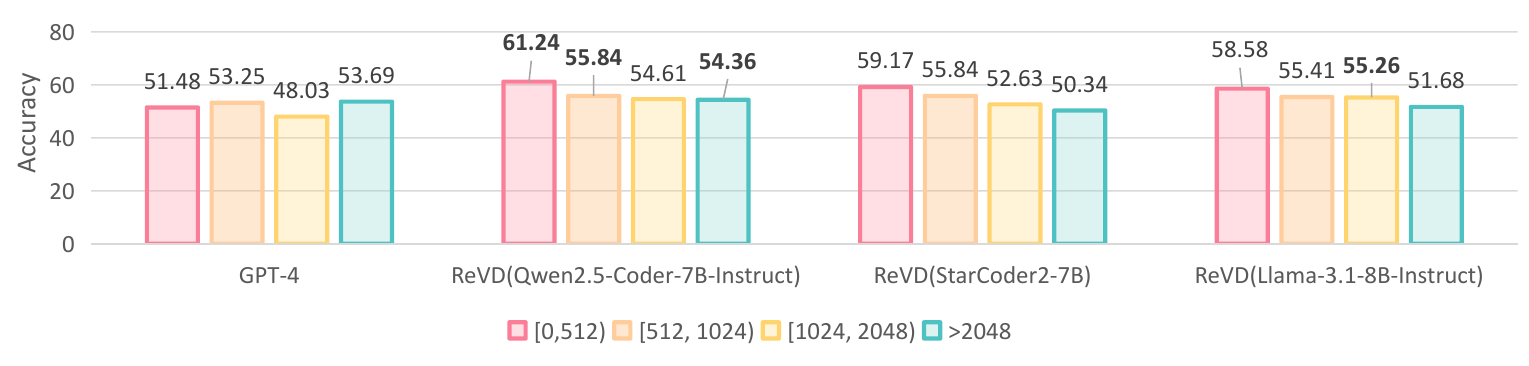}
    \caption{\textbf{The accuracy of \tool under varying token length.} The \textcolor{mypink}{pink},  \textcolor{myorange}{orange}, \textcolor{myyellow}{yellow}, and \textcolor{mygreen}{green} denote the \tool performance for token numbers containing \textcolor{mypink}{less than 512 tokens}, \textcolor{myorange}{between 512 and 1024 tokens}, \textcolor{myyellow}{between 1024 and 2048 tokens}, and \textcolor{mygreen}{more than 2048 tokens}.
   } 

\label{appendix_figure_rq}
\end{figure*}

In this section, as shown in Table~\ref{appendix_figure_rq}, we show the complete experiment results in different token lengths of \tool involved with Qwen2.5-Coder-7B-Instruct, Llama-3.1-8B-Instruct and StarCoder2-7B, respectively.

\section{Prompts for Vulnerability Detection}
In this section, we describe prompts of LLMs to detect vulnerabilities. We utilize the prompting technique developed by Ding et al~\cite{DBLP:journals/corr/abs-2403-18624/primevul}. For practical implementation, we configure the system to accept a maximum input length of 8192 tokens and allow a maximum of 2048 tokens to be generated for each code snippet.

\begin{table}[htb]
\scriptsize
\begin{minipage}{0.9\linewidth}
The prompt of LLMs to detect vulnerabilities.
\centering
\ttfamily
\begin{tabular}{p\linewidth}
\midrule
\textbf{[System Prompt]:}  \\
You are a security expert that is good at static program analysis.\\
\textbf{[User Prompt]:}  \\
""" \\
Please analyze the following code:  \\
```\\
{func}\\
```\\
Please indicate your analysis result with one of the options: \\
(1) YES: A security vulnerability detected.\\
(2) NO: No security vulnerability. \\
Make sure to include one of the options above "explicitly" \\(EXPLICITLY!!!) in your response.\\
Let's think step-by-step.\\
"""
\\
\midrule
\end{tabular}
\end{minipage}%
\label{tab:answer:GPT4}
\end{table}

\section{Complete Answer in Figure~\ref{example}}
\label{appendix:answer}
In this section, we present a comprehensive analysis of the responses provided by GPT-4 and \tool for the example~\cite{CVE-2021-37683} illustrated in Figure~\ref{example}. Our findings indicate that GPT-4 fails to accurately distinguish between the pre-code, which is vulnerable, and the post-code, which is non-vulnerable. Conversely, \tool not only accurately identifies the pre-code as vulnerable and the post-code as non-vulnerable but also specifies the type of vulnerability, the mechanisms triggering the vulnerability, and the underlying causes. This detailed assessment underscores the superior diagnostic capabilities of \tool in identifying and analyzing software vulnerabilities.

\begin{table*}[htb]
\scriptsize
\begin{minipage}{0.9\linewidth}
The complete answer of pre-code and post-code by GPT-4 in Figure~\ref{example}.
\centering
\ttfamily
\begin{tabular}{p\linewidth}
\midrule
\textbf{For Pre-code Answer:}  \\
The provided code is a TensorFlow Lite function for evaluating a division operation. It takes two input tensors, performs a division operation, and stores the result in an output tensor. The function supports different data types for the output tensor, including floating point, integer, and quantized unsigned integer.\\ \\The function begins by obtaining the parameters and data for the operation from the node's built-in and user data. It then retrieves the input and output tensors. If the output tensor's type is either floating point or integer, it calls the `EvalDiv` function. If the output tensor's type is a quantized unsigned integer, it calls the `EvalQuantized` function. If the output tensor's type is not supported, it reports an error and returns a failure status.\\ \\From a security perspective, the function seems to be safe. It checks the return status of all function calls that could potentially fail, and it handles unsupported output tensor types gracefully. It does not perform any unsafe memory operations, and it does not seem to be susceptible to common security vulnerabilities such as buffer overflows, integer overflows, or use-after-free vulnerabilities.\\ \\Therefore, my analysis result is: \\ \\(2) NO: No security vulnerability.

\\
\textbf{For Post-code Answer:} \\
The provided code is a part of TensorFlow Lite, which is a set of tools to help developers run TensorFlow models on mobile, embedded, and IoT devices. It enables on-device machine learning inference with low latency and a small binary size.\\ \\
The function `Eval` is used to evaluate a division operation on two tensors. It checks the type of the output tensor and performs the division operation accordingly. For `kTfLiteFloat32` type, it directly performs the division operation. For `kTfLiteInt32` and `kTfLiteUInt8` types, it first checks if the divisor is non-zero to avoid division by zero error. If the output type is not one of these three, it reports an error.\\ \\
From a security perspective, the code seems to be safe. It checks for division by zero for integer types, which could otherwise lead to undefined behavior. It also checks the validity of the input and output tensors before performing operations on them. If any of these checks fail, it returns an error status. \\ \\
However, the code assumes that the `node` pointer passed to the function is not null and that the `builtin\_data` and `user\_data` fields of the `node` structure are valid pointers. If this is not the case, the code could lead to a segmentation fault. But this is more of a robustness issue than a security vulnerability, as exploiting it would require control over the internal state of the program, which is not typically possible for an external attacker.\\ \\
Therefore, my analysis result is: \\ \\
(2) NO: No security vulnerability. \\ \\
However, it's recommended to add null pointer checks for `node`, `node->builtin\_data`, and `node->user\_data` to improve the robustness of the code.\\
\midrule
\end{tabular}
\end{minipage}%
\label{tab:answer:GPT4}
\end{table*}

\begin{table*}[htb]
\scriptsize
\begin{minipage}{0.9\linewidth}
The complete answer of pre-code and post-code by \tool in Figure~\ref{example}.
\centering
\ttfamily
\begin{tabular}{p\linewidth}
\midrule
\textbf{For Pre-code Answer:}  \\
YES: A security vulnerability detected.\\
Reason:\\
$[$Interpretation$]$:\\
1. The vulnerability in target code arises in the lack of a check for empty input tensors before the division operation is performed. Specifically, line 16 of the target code does not check if the input tensors are empty before proceeding with the division.\\
2. Specifically, line 16 of the target code contains the `\texttt{EvalDiv<kernel\_type>}` function call, which is executed if the output tensor type is either `\texttt{kTfLiteFloat32}` or `\texttt{kTfLiteInt32}`. The control flow includes checking the type of the output tensor and then calling the appropriate evaluation function. The data flow includes passing the context, node, parameters, data, and input tensors to the evaluation function.\\
3. First, the vulnerability in the division operation arises from the lack of a check for empty input tensors before the division is performed. Then, in combination with the control flow that directly proceeds to the division evaluation without any checks and the data flow that passes the input tensors directly to the division function, this can lead to a division by zero error if the input tensors are empty. Finally, this target code highlights the importance of validating input data sizes before performing operations that assume non-empty input data.

\\
\textbf{For Post-code Answer:} \\
NO: No security vulnerability.\\
Reason:\\
$[$Interpretation$]$:\\
1. The fixed code in target code arises in lines 23-25 and 31-33.\\
2. Specifically, line 23 contains the macro `\texttt{TF\_LITE\_CHECK\_DIV\_NON\_ZERO(data\_type)}` which is used to check if the division operation is supported for a given data type. The control flow contains a check for the output tensor type and then a conditional check for the input tensor type. The data flow contains the input tensors being validated before they are used in the division operation.
\\3. First, the fixed reason in line 23 arises from the removal of the division by zero check for the `\texttt{kTfLiteFloat32}` output tensor type. Then, in combination with the control flow in lines 23-25 and the data flow in lines 31-33, the vulnerability is mitigated because the check for division by zero is now only performed for the `\texttt{kTfLiteInt32}` and `\texttt{kTfLiteUInt8}` output tensor types, and even then, it is only a debug assertion for `\texttt{kTfLiteUInt8}`. Finally, this target code highlights that the vulnerability does not trigger because the necessary checks are in place to prevent division by zero for the supported tensor types.\\
\midrule
\end{tabular}
\end{minipage}%
\label{tab:answer:tool}
\end{table*}

\section{Manual Analysis}
We conducted a careful examination of our generative reasoning data. Due to time constraints and heavy human costs, we randomly selected 20 examples for manual vulnerability analysis. This analysis was independently performed by two developers, each with over five years of experience in the field of vulnerability security. If two developers have differing opinions on the same reasoning answer, the two experts will discuss with each other to reach a consensus. Finally, our dataset exhibits higher consistency than those derived from the NVD, highlighting the robustness of our data generation process (95\% accuracy). We also provided a detailed case included in the appendix, where the BVD accurately identifies specific types of vulnerabilities, providing insights into the program's data flow and control flow, and specifying the triggers involved. More examples will be updated in our open-source repository.


\end{document}